# Lightweight Stochastic Video Prediction via Hybrid Warping

Kazuki Kotoyori[†], Shota Hirose[†], Heming Sun[‡], Jiro Katto[†]

[†]School of Fundamental Science and Engineering, Waseda University, Tokyo, Japan

[‡]Faculty of Engineering, Yokohama National University, Kanagawa, Japan

kotokoto.kaz@toki.waseda.jp, syouta.hrs@akane.waseda.jp, sun-heming-vg@ynu.ac.jp, katto@waseda.jp

*Abstract*— Accurate video prediction by deep neural networks, especially for dynamic regions, is a challenging task in computer vision for critical applications such as autonomous driving, remote working, and telemedicine. Due to inherent uncertainties, existing prediction models often struggle with the complexity of motion dynamics and occlusions. In this paper, we propose a novel stochastic long-term video prediction model that focuses on dynamic regions by employing a hybrid warping strategy. By integrating frames generated through forward and backward warpings, our approach effectively compensates for the weaknesses of each technique, improving the prediction accuracy and realism of moving regions in videos while also addressing uncertainty by making stochastic predictions that account for various motions. Furthermore, considering real-time predictions, we introduce a MobileNet-based lightweight architecture into our model. Our model, called SVPHW, achieves state-of-the-art performance on two benchmark datasets.

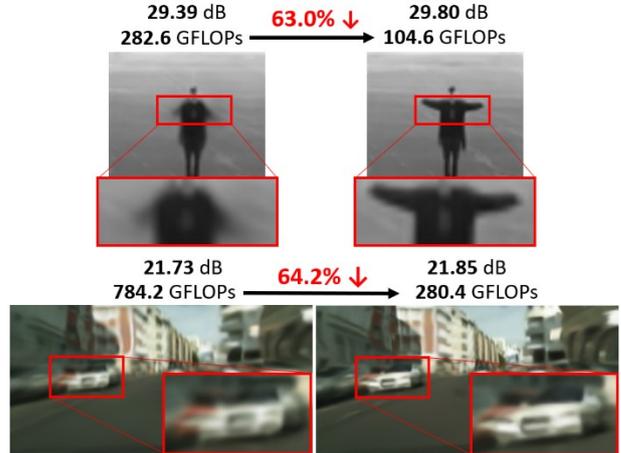

Fig. 1. Visal comparison with PSNR [dB] and computational costs (GFLOPs) of future frames generated by SLAMP (**left**) [10] and SVPHW (**right**) on KTH [14] (**top**) and Cityscapes [15] (**bottom**) datasets, where KTH frames are the 20th and Cityscapes the 10th prediction frames.

## I. Introduction

In video prediction, a machine learning-based model predicts future frames from a video by referring to past frames. This task has gained a lot of attention, due to the numerous possible application such as autonomous driving [1] and low-latency video transmission systems [2, 3]. Given the high level of safety and usability of these systems, it is essential to predict future frames accurately and quickly. However, it is difficult to achieve both simultaneously. In addition, it is more difficult to predict long-term video frames than short-term ones because of the error accumulation and increased uncertainty. Various approaches have recently been proposed to address these difficulties, such as low-computational-cost models [4, 5, 6], stochastic prediction models [7, 8, 9, 10, 11], and optical flow-based models [4, 5, 6, 10, 11, 12].

Optical flow-based models generate frames by warping using an optical flow that represents the movement of each pixel between frames. These models often use backward warping to generate future frames because it is less difficult to implement and can generate higher-quality frames than those generated by forward warping. However, backward warping is a sampling task and cannot use all pixels of the reference frame, as shown in Fig. 2. This causes the problem of missing information regarding the reference frame. On the other hand, forward warping is a splatting task; therefore, it can use all the pixels from the reference frame, but it also has the problem of multiple pixel mapping and holes, as shown in Fig. 2.

We introduce hybrid warping, using forward and backward optical flows, into a stochastic prediction model. This model is known as Lightweight Stochastic Video Prediction via Hybrid Warping (SVPHW). Using forward and backward warpings may utilize each strength and alleviate each weakness. In addition, we used appearance-specific frames [10] to inpaint occluded regions that are difficult to predict with warping. When generating these two warped frames and appearance frames, three separately estimated stochastic latent variables were input into the model, in addition to past frames. This allows the model to predict a wide variety of motion in videos as shown in Fig. 2 and thus cope with long-term predictions with high uncertainty. To reduce the computational cost, we also introduced a MobileNet-based lightweight architecture [13], which is proposed for image recognition, into the encoders and decoders. Through experiments, we show that the SVPHW is a high-performance prediction model compared to existing models in terms of prediction accuracy and the computational cost. In addition, a visual comparison of the frames generated by each prediction model shows that the SVPHW can predict dynamic parts in videos (see Fig. 1).

Our contributions are summarized as follows.

- We proposed a hybrid warping method for stochastic video prediction. This approach improves the long-term and accurate prediction of dynamic parts in videos.

- We introduced a MobileNet-based architecture into the encoders and decoders of an SVPHW. The SVPHW significantly reduces the computational cost using this architecture.

- Experiments show that the SVPHW achieves the highest prediction accuracy compared to existing models in objective and subjective evaluations at the lowest computational cost.

## II. Related Work

### A. Video Prediction

In recent years, a wide range of video prediction techniques have been proposed [16]. PredNet [17] trained an unsupervised neural network and used a convolutional LSTM [18] to retain time-series dependencies and spatial information. OPT [12], DMVFN [4] and IFRVP [6] generate future frames by backward warping and only use two past frames to predict a future frame. These models performed well in short-term predictions, and DMVFN and IFRVP performed particularly well at a low computational cost.

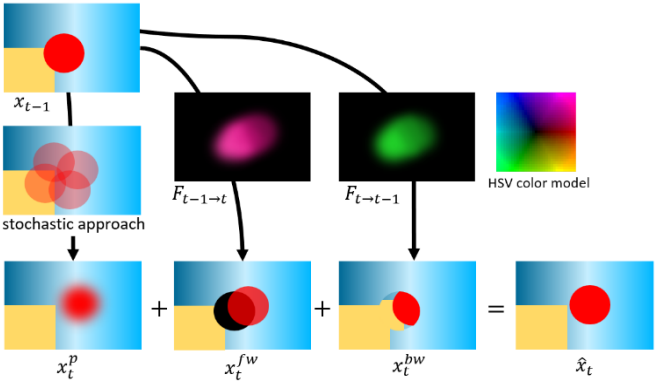

Fig. 2. Example of generating a future frame $\hat{x}_t$ using hybrid warping, referring to a frame $x_{t-1}$ and optical flows. $x_t^p$ represents an appearance-specific frame with stochastic approach, $x_t^{fw}$ and $x_t^{bw}$ represent frames by forward and backward warpings, $F_{t-1 \to t}$ and $F_{t \to t-1}$ represent forward and backward optical flows. $F_{t-1 \to t}$ and $F_{t \to t-1}$ are also gained with stochastic approach.

Although the aforementioned models are deterministic methods, numerous stochastic prediction models have been proposed to predict various motion patterns in videos. SV2P [7] and SVG [8] are the first stochastic prediction models. In the stochastic approach, past frames create a posterior distribution for predicting future frames. In this context, learning involves maximizing the likelihood of the observed data while minimizing the divergence between the prior and posterior distributions. Among the stochastic methods, SLAMP [10] is known as a model with particularly high prediction accuracy, using LSTMs, which are effective for long-term video prediction. SLAMP is also effective for videos with a large moving range using optical flow. However, SLAMP uses multiple encoders and decoders for the prediction task, resulting in a larger model.

Other approaches have also been proposed, such as MCVD [19] and ExtDM [5], which incorporate diffusion models and have recently attracted attention in image generation. In particular, ExtDM utilizes latent space and is faster than other diffusion models while maintaining high prediction accuracy.

### B. MobileNet

A lightweight architecture is an important factor in the development of image-processing-based applications. MobileNet [13] is an efficient neural network architecture designed for mobile and embedded vision applications. The main innovation is the use of depthwise separable convolution, which significantly reduces the number of parameters and computational complexity. Depthwise separable convolution splits the standard convolution into a depthwise convolution, which applies one filter per input channel, and a pointwise convolution, which uses a 1×1 convolution to combine outputs. This architecture is lightweight although its performance is slightly lower than before splitting. Improved versions of MobileNetv2 [20] and MobileNetv3 [21] were later proposed. The former further reduces the computational complexity of pointwise convolution by inverted residuals based on ResNet [22], whereas the latter further improves the accuracy by introducing squeeze-and-excitation (SENet) [23] and a ReLU6-based activation function called h-swish. However, the simplest structure [13] has the lowest computational cost compared to the improved versions [20, 21] when the conditions for the number of input and output channels are the same.

### III. PROPOSED METHOD

#### A. Stochastic Video Prediction via Hybrid Warping

Our proposed method, called SVPHW, was inspired by the SLAMP [10] architecture. In the stochastic video prediction method, we used past frames $x_{1:t-1}$ and stochastic latent variables $z_{1:t}$ to predict future frames $x_t$. For prediction, the recurrent model was used, with only $x_{t-1}$ and $z_t$ as inputs. In the training process, the posterior distribution $q_\phi(z_t|x_{1:t})$ was computed from all past frames, including the target frame to be predicted, and the stochastic latent variable $z_t$ was sampled from that distribution. The model was trained to reduce the distance between this posterior distribution $q_\phi(z_t|x_{1:t})$ and the prior distribution $p_\psi(z_t|x_{1:t-1})$ through KL-divergence.

The SVPHW components are shown in Fig. 3 (a). SVPHW generates three types of future frames $x_t^p$, $x_t^{fw}$, and $x_t^{bw}$, which are fused for the final prediction $\hat{x}_t$ by their weighted sum. $x_t^p$ specializes in appearance, such as in occluded regions. The latest frame $x_{t-1}$ was encoded in the pixel encoder. Then, from the output appearance features and the latent variable $z_t^p$, the temporal and spatial features were captured using a convolutional LSTM to predict the next frame. Finally, $x_t^p$ was obtained using a pixel decoder. For the motion-specific $x_t^{fw}$ and $x_t^{bw}$, the two latest frames $x_{t-2}$ and $x_{t-1}$ were input to the forward and backward motion encoders, respectively, to extract the motion information between frames. Convolutional LSTMs were also used with the latent variables $z_t^{fw}$ and $z_t^{bw}$. Forward and backward flow decoders output the next forward and backward optical flows $F_{t-1 \to t}$ and $F_{t \to t-1}$, respectively. Subsequently, $x_t^{fw}$ and $x_t^{bw}$ were generated through forward and backward warpings. The generation of the $x_t^p$ and $x_t^{bw}$ process was the same as that in the SLAMP method. However, forward warping often caused problems, such as multiple pixel mapping and holes. To alleviate these problems, we applied a forward warping technique called average splatting [24], which averages the pixel values around the occluded regions to generate $x_t^{fw}$. Finally, $x_t^p$, $x_t^{fw}$, and $x_t^{bw}$ were input to the mask decoder, and weight maps $m^p$, $m^{fw}$, and $m^{bw}$ were output. The final prediction $\hat{x}_t$ is expressed as follows:

$$\hat{x}_t = m^p \odot x_t^p + m^{fw} \odot x_t^{fw} + m^{bw} \odot x_t^{bw}, \quad (1)$$
$$m^p + m^{fw} + m^{bw} = 1, \quad m^p, m^{fw}, m^{bw} \in [0,1]$$

where $\odot$ denotes element-wise Hadamard product.

Fig. 3 (c) shows stochastic latent variable predictors. This figure represents the test process; therefore, the latent variables were obtained from the prior distributions. We approximated the posterior distributions $q_{\phi_p}(z_t^p|x_{1:t})$, $q_{\phi_{fw}}(z_t^{fw}|x_{1:t})$, and $q_{\phi_{bw}}(z_t^{bw}|x_{1:t})$ by a conditional Gaussian distribution to train the distribution of the latent variables $z_t^p$, $z_t^{fw}$, and $z_t^{bw}$. We trained the model to reduce the distance between the posterior and prior distributions by KL divergence; however, because $z_t^p$, $z_t^{fw}$, and $z_t^{bw}$ are

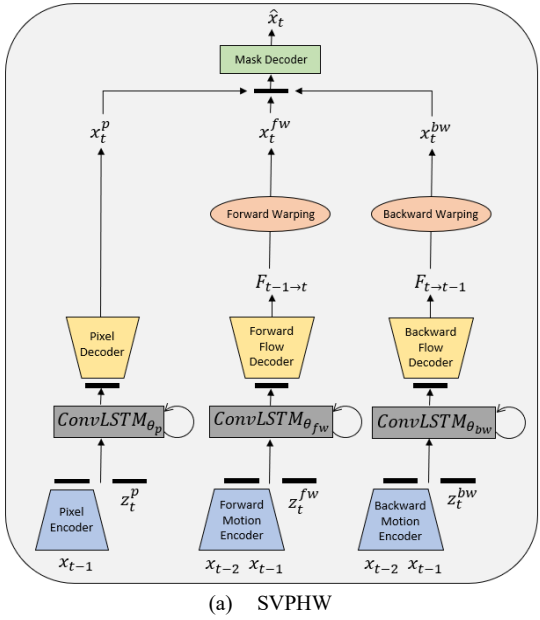
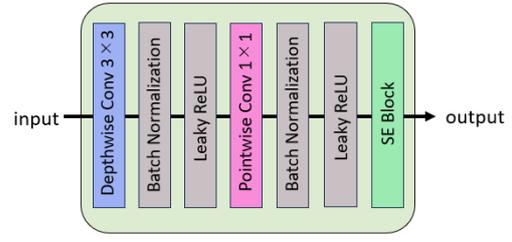
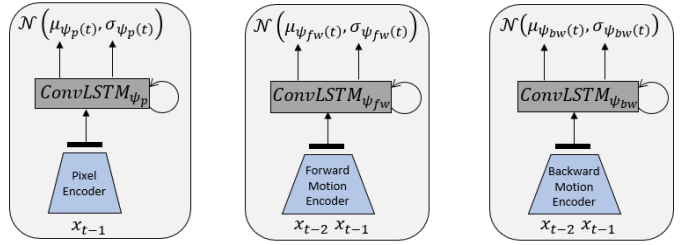

Fig. 3. Overview of proposed architectures. (a) SVPHW components, (b) lightweight MobileNet with Squeeze-and-Excitation (MNSE) layer, and (c) stochastic latent variable predictors components.

independent in time, we trained it to decompose the KL terms and optimize the variational lower bound:

$$\mathcal{L}_{\theta,\phi_p,\phi_{fw},\phi_{bw},\psi_p,\psi_{fw},\psi_{bw}}(x_{1:T}) \quad (2)$$

$$= \sum_t \mathbb{E}_{\substack{z^p_{1:t} \sim q_{\phi_p} \\ z^{fw}_{1:t} \sim q_{\phi_{fw}} \\ z^{bw}_{1:t} \sim q_{\phi_{bw}}}} \log p_\theta(x_t | x_{1:t-1}, z^p_{1:t}, z^{fw}_{1:t}, z^{bw}_{1:t})$$

$$-\beta \left[ D_{KL}\left(q_{\phi_p}\left(z^p_t|x_{1:t}\right) || p_{\psi_p}\left(z^p_t|x_{1:t-1}\right)\right) \right.$$

$$+ D_{KL}\left(q_{\phi_{fw}}\left(z^{fw}_t|x_{1:t}\right) || p_{\psi_{fw}}\left(z^{fw}_t|x_{1:t-1}\right)\right)$$

$$\left. + D_{KL}\left(q_{\phi_{bw}}(z^{bw}_t|x_{1:t}) || p_{\psi_{bw}}(z^{bw}_t|x_{1:t-1})\right) \right]$$

The likelihood $p_\theta$ can be regarded as the sum of the $L_2$ penalty between the target frame $x_t$ and each generated frame $x^p_t$, $x^{fw}_t$, $x^{bw}_t$ and $\hat{x}_t$. $\beta$ is weight for KL terms.

### B. MobileNet with Squeeze-and-Excitation

As SVPHW uses more encoders and decoders than SLAMP [10], its computational cost is higher. To cope with this, we used a MobileNet-based architecture [13] called MobileNet with Squeeze-and-Excitation (MNSE), as shown in Fig. 3 (b). By introducing a MobileNet-based architecture, we reduced the computational cost by using $(C_{out} + K^2)/C_{out}K^2$ FLOPs over the standard convolution in SLAMP, where $C_{out}$ is the number of output channels and $K$ is the kernel size. MobileNet has the lowest computational cost among MobileNetv2 [20] and MobileNetv3 [21] for the same number of input and output channels, respectively. However, the accuracy was lower than that of these two methods. This problem was addressed by adding a squeeze-and-excitation (SE) block [23] that adaptively weighs each channel of the convolution layer, instead of outputting it equally. This increased the expressiveness of the model and improved the prediction accuracy with a slight increase in the computational cost.

## IV. EXPERIMENTS

### A. Experimental Conditions

We used two datasets to train and evaluate the SVPHW: KTH [14] (64×64), a dataset of human beings filmed by a fixed camera performing single actions such as running or hand-weaving, and Cityscapes [15] (128×256), a dataset of car frontal footage filmed in 50 European cities. In the training process, we conditioned 10 frames and predicted 10 future frames on both datasets; however, the number of epochs was 300 for KTH and 150 for Cityscapes. In terms of GPU memory limitations and a further reduction in the computational cost, we reduced the number of input and output channels in the MNSE layer to 3/4 of the standard convolution in SLAMP for KTH and 1/4 for Cityscapes. During testing, we also conditioned 10 frames on both datasets; however, we predicted 30 frames on KTH and 20 frames on Cityscapes. These experimental conditions follow SLAMP [10]. To objectively evaluate the prediction accuracy of our model, we used two metrics: PSNR and SSIM [25]. We also calculated GFLOPs and the number of parameters to measure the computational cost.

### B. Quantitative and Qualitative Results

We compared the SVPHW with state-of-the-art methods. The quantitative results for KTH [14] are listed in Table I. PSNR and SSIM are averages over all test sequences, and the values of the existing models are taken from the original papers, whereas GFLOPs were measured when generating one future frame. The table shows that the SVPHW achieves the best PSNR 29.80 dB and the lowest computational cost 104.6 GFLOPs compared with the other models. Fig. 5 shows the qualitative results of SVPHW. The frames t=1 to t=10 in the first row are conditioning frames for the model to use for prediction and the weight maps indicate that the white areas were strongly weighted. Notably, in the warped frames of Fig. 5, $x^{fw}_t$ and $x^{bw}_t$ have a higher contribution to the moving object or its edge, whereas $x^p_t$ has a higher contribution to the occluded regions around the moving part. These results indicate that the appearance prediction and hybrid-warped frames work together to achieve highly accurate predictions.

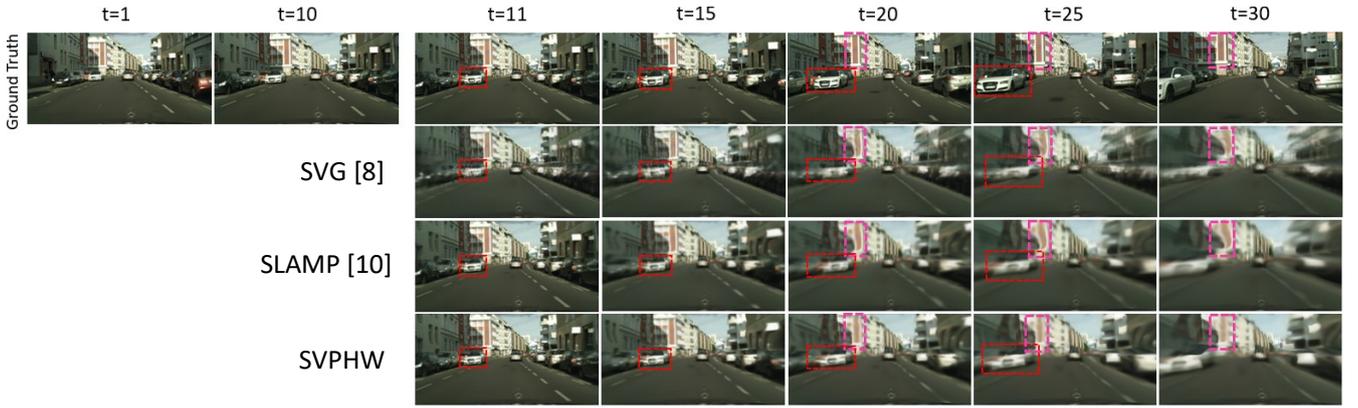

Fig. 4. Qualitative results of long-term prediction models on Cityscapes [15] (128×256). Top row shows ground truth and conditioning frames up to t=10, followed by long-term prediction models.

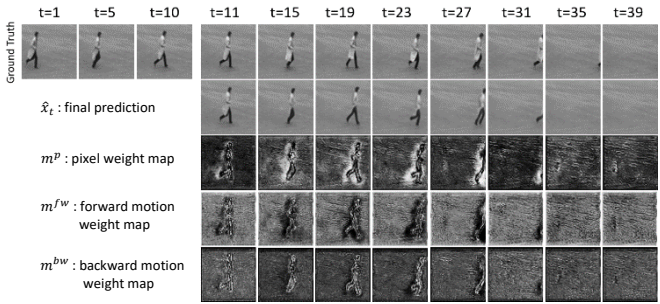

Fig. 5. Qualitative results of SVPHW on KTH [14] (64×64). The top row shows the ground truth and conditioning frames up to t=10, followed by final prediction and three weight maps. The weight maps are weighted more strongly in white areas.

The quantitative results for the Cityscapes [15] are listed in Table II. The table shows that our PSNR and SSIM are 21.85 dB and 0.654, which are the highest of all. Also, our computational cost is 280.4 GFLOPs, which is the lowest of all. Fig. 1 and 4 are the qualitative results and show that SVPHW preserves the shape of the background and moving objects such as cars. These results indicate that the SVPHW is particularly effective in predicting the future of images with a wide range of motion, such as driving datasets, owing to hybrid warping. In addition, the results for KTH and Cityscapes show that even if the number of input and output channels in each encoder and decoder layer is reduced, a high prediction accuracy can be maintained while reducing the computational cost owing to the MNSE layer.

Table III shows the number of parameters in encoders and decoders. We should note that additional image encoder is introduced for Cityscapes before each encoder in Fig. 3 to reduce image sizes similar to SLAMP [10]. The table shows that SVPHW has considerably fewer encoders and decoders parameters than others. In particular, on the Cityscapes dataset, the number of SVPHW parameters are reduced to about 1/40 of SLAMP. This result indicate SVPHW is a very lightweight model owing to the MNSE layer.

## V. Conclusion

In this study, we propose Lightweight Stochastic Video Prediction via Hybrid Warping (SVPHW), which combines forward- and backward-warped frames with a stochastic method to predict the dynamic parts in future frames. Furthermore, we introduced MobileNet with Squeeze-and-

TABLE I. Quantitative results and GFLOPs of long-term prediction models for KTH [14].

| Models | KTH [14] (64×64) | | |
|---|---|---|---|
| | PSNR (↑) | SSIM (↑) | GFLOPs |
| SVG [8] | 28.06 | 0.844 | 128.9 |
| SLAMP [10] | 29.39 | **0.865** | 282.6 |
| MCVD [19] | 27.50 | 0.835 | 28855.0 |
| ExtDM-K3 [5] | 29.04 | 0.817 | 2523.1 |
| SVPHW | **29.80** | 0.863 | **104.6** |

TABLE II. Quantitative results and GFLOPs of long-term prediction models for Cityscapes [15].

| Models | Cityscapes [15] (128×256) | | |
|---|---|---|---|
| | PSNR (↑) | SSIM (↑) | GFLOPs |
| SVG [8] | 20.42 | 0.606 | 336.4 |
| SLAMP [10] | 21.73 | 0.649 | 784.2 |
| SVPHW | **21.85** | **0.654** | **280.4** |

TABLE III. The number of parameters in encoders and decoders of stochastic long-term prediction models.

| Models | Million Parameters | |
|---|---|---|
| | KTH [14] | Cityscapes [15] |
| SVG [8] | 20.50 | 10.37 |
| SLAMP [10] | 41.12 | 17.62 |
| SVPHW | **3.43** | **0.43** |

Excitation (MNSE) in encoders and decoders to reduce the computational cost during the prediction task. Through experiments, the SVPHW achieved state-of-the-art performance at the lowest computational cost. In future work, we will continue to reduce the computational cost while preserving high prediction accuracy, such as under 100 or 10 GFLOPs.

## VI. Acknowledgement

This paper is supported by the Ministry of Internal Affairs and Communications's project for efficient frequency utilization toward wireless IP multicasting.

## References

[1] A. Bhattacharyya, M. Fritz, and B. Schiele, "Long-term on-board prediction of people in traffic scenes under uncertainty," *2018 IEEE/CVF Conference on Computer Vision and Pattern Recognition (CVPR)*, Salt Lake City, UT, USA, 2018, pp. 4194-4202. doi: 10.1109/CVPR.2018.00441

[2] Y. Katsuyama *et al*., "A predictive approach for compensating transmission latency in remote robot control for improving teleoperation efficiency," *GLOBECOM 2023 - 2023 IEEE Global Communications Conference*, Kuala Lumpur, Malaysia, 2023, pp. 6934-6939, doi: 10.1109/GLOBECOM54140.2023.10437076.

[3] T. Sato *et al*., "Compensation of communication latency using video prediction in remote monitoring systems," *2023 International Conference on Emerging Technologies for Communications (ICETC 2023)*, 2023.

[4] X. Hu, Z. Huang, A. Huang, J. Xu and S. Zhou, "A dynamic multi-scale voxel flow network for video prediction," *2023 IEEE/CVF Conference on Computer Vision and Pattern Recognition (CVPR)*, Vancouver, BC, Canada, 2023 pp. 6121-6131. doi: 10.1109/CVPR52729.2023.00593

[5] Z. Zhang, J. Hu, W. Cheng, D. Paudel, and J. Yang, "Extdm: Distribution extrapolation diffusion model for video prediction," *IEEE/CVF Conference on Computer Vision and Pattern Recognition (CVPR)*, 2024, pp. 19310-19320.

[6] S. Hirose, K. Kotoyori *et al*., "Real-time video prediction with fast video interpolation model and prediction training," *2024 IEEE Internationak Conference on Image Processing (ICIP)*, 2024.

[7] M. Babaeizadeh, C. Finn, D. Erhan, R. H. Campbell, and S. Levine, "Stochastic variational video prediction," *Proc. of the International Conf. on Learning Representations (ICLR),* 2018.

[8] E. Denton and R. Fergus, "Stochastic video generation with a learned prior," *Proceedings of the 35th International Conference on Machine Learning (PMLR)*, 2018, pp. 1174-1183.

[9] A. X. Lee *et al*., "Stochastic adversarial video prediction," 2018, arXiv:1804.01523. [Online]. Available: https://doi.org/10.48550/arXiv.1804.01523

[10] A. K. Akan, E. Erdem, A. Erdem, and F. Gunrey, "Slamp: Stochastic latent appearance and motion prediction," *Proceedings of the IEEE/CVF International Conference on Computer Vision (ICCV)*, 2021, pp. 14728-14737.

[11] A. K. Akan, S. Safadoust, and F. Guney, "Stochastic video prediction with structure and motion," 2022, arXiv:2203.10528. [Online]. Available: https://doi.org/10.48550/arXiv.2203.10528

[12] Y. Wu, Q. Wen, and Q. Chen, "Optimizing video prediction via video frame interpolation," *2022 IEEE/CVF Conference on Computer Vision and Pattern Recognition (CVPR),* New Orleans, LA, USA, 2022, pp. 17793-17802. doi: 10.1109/CVPR52688.2022.01729

[13] A. G. Howard *et al*., "Mobilenets: Efficient convolutional neural networks for mobile vision applications," 2017, arXiv:1704.04861, [Online]. Available: https://doi.org/10.48550/arXiv.1704.04861

[14] C. Schüldt, I. Laptev, and B. Caputo, "Recognizing human actions: a local SVM approach," *Proceedings of the 17th International Conference on Pattern Recognition, 2004. ICPR 2004.*, Cambridge, UK, 2004, pp. 32-36 Vol.3, doi: 10.1109/ICPR.2004.1334462.

[15] M. Cordts *et al*., "The cityscapes dataset for semantic urban scene understanding," *2016 IEEE Conference on Computer Vision and Pattern Recognition (CVPR)*, Las Vegas, NV, USA, 2016, pp. 3213-3223, doi: 10.1109/CVPR.2016.350.

[16] S. Oprea *et al*., "A review on deep learning techniques for video prediction," *IEEE Transactions on Pattern Analysis and Machine Intelligence*, vol. 44, no. 6, pp. 2806-2826, 1 June 2022, doi: 10.1109/TPAMI.2020.3045007.

[17] W. Lotter, G. Kreiman, and D. Cox, "Deep predictive coding networks for video prediction and unsupervised learning," *International Conference on Learning Representations (ICLR)*, 2017

[18] X,Shi, Z. Chen *et al*., "Convolutional LSTM network: A machine learning approach for precipitation nowcasting," *Advances in Neural Information Processing Systems (NeurIPS)*, pp. 802-810, 2015.

[19] V. Voleti, A. Jolicoeur-Martineau, and C. Pal, "MCVD: Masked conditional video diffusion for prediction, generation, and interpolation," *Advances in Neural Information Processing Systems (NeurIPS)*, pp. 23371–23385, 2022.

[20] S. Mark, A. Howard, M. Zhu, A. Zhmoginov and L. Chen, "MobileNetV2: Inverted residuals and linear bottlenecks," *2018 IEEE/CVF Conference on Computer Vision and Pattern Recognition (CVPR)*, Salt Lake City, UT, USA, 2018, pp. 4510-4520. doi: 10.1109/CVPR.2018.00474

[21] A. Howard *et al*., "Searching for MobileNetV3," *2019 IEEE/CVF International Conference on Computer Vision (ICCV)*, Seoul, Korea (South), 2019, pp. 1314-1324, doi: 10.1109/ICCV.2019.00140.

[22] K. He, X. Zhang, S. Ren and J. Sun, "Deep residual learning for image recognition," *2016 IEEE Conference on Computer Vision and Pattern Recognition (CVPR)*, Las Vegas, NV, USA, 2016, pp. 770-778, doi: 10.1109/CVPR.2016.90.

[23] J. Hu, L. Shen, and G. Shun, "Squeeze-and-excitation networks," *Proceedings of the IEEE Conference on Computer Vision and Pattern Recognition (CVPR),* 2018, pp. 7132-7141

[24] S. Niklaus and F. Liu, "Softmax splatting for video frame interpolation," *Proceedings of the IEEE/CVF Conference on Computer Vision and Pattern Recognition (CVPR)*, 2020, pp. 5437-5446

[25] Z. Wang, A. C. Bovik, H. R. Sheikh and E. P. Simoncelli, "Image quality assessment: from error visibility to structural similarity," *IEEE Transactions on Image Processing*, vol. 13, no. 4, pp. 600-612, April 2004, doi: 10.1109/TIP.2003.819861.